\documentclass[twoside,11pt]{article}

\usepackage{cite}
\usepackage{amsmath,amssymb,amsfonts}
\usepackage{algorithmic}
\usepackage{graphicx}
\usepackage[toc,page]{appendix}
\usepackage{subfigure}
\usepackage{textcomp}
\usepackage{xcolor}
\usepackage{jmlr2e}
\begin{document}

\title{CAMTA: Causal Attention Model for Multi-touch Attribution}

\author{\name Sachin Kumar \email sachink382@iitkgp.ac.in \\
       \addr Indian Institute of Technology, Kharagpur\\
       \AND
       \name Garima Gupta \email gupta.garima1@tcs.com \\
       \addr TCS Research \& Innovation, Delhi\\
       \AND
       \name Ranjitha Prasad \email ranjitha@iiit.ac.in\\
       \addr IIIT-Delhi\\
       \AND
       \name Arnab Chatterjee \email arnab.chatterjee4@tcs.com \\
       \addr TCS Research \& Innovation, Delhi\\
       \AND
       \name Lovekesh Vig \email lovekesh.vig@tcs.com \\
       \addr TCS Research \& Innovation, Delhi\\
       \AND 
       \name Gautam Shroff \email gautam.shroff@tcs.com\\
       \addr TCS Research \& Innovation, Delhi\\}

%\editor{Kevin Murphy and Bernhard Sch{\"o}lkopf}

\maketitle

\begin{abstract}%   <- trailing '%' for backward compatibility of .sty file
Advertising channels have evolved from conventional print media, billboards and radio-advertising to online digital  advertising (ad), where the users are exposed to a sequence of ad campaigns via social networks, display ads, search etc. While advertisers revisit the design of ad campaigns to concurrently serve the requirements emerging out of new ad channels, it is also critical for advertisers to estimate the contribution from touch-points (view, clicks, converts) on different channels, based on the sequence of customer actions. This process of contribution measurement is often referred to as multi-touch attribution (MTA). In this work, we propose \texttt{CAMTA}\footnote{Published in Data Mining for Service Workshop ICDM 2020}, a novel deep recurrent neural network architecture which is a causal attribution mechanism for user-personalised MTA in the context of observational data. \texttt{CAMTA} minimizes the selection bias in channel assignment across time-steps and touchpoints. Furthermore, it utilizes the users’ pre-conversion actions in a principled way in order to predict per-channel attribution. To quantitatively benchmark the proposed MTA model, we employ the real-world Criteo dataset and demonstrate the superior performance of \texttt{CAMTA} with respect to prediction accuracy as compared to several baselines. In addition, we provide results for budget allocation and user-behaviour modeling on the predicted channel attribution.
\end{abstract}

\begin{keywords}
Multi-touch attribution, Causality, temporal modeling, Touchpoints, Attention, Conversion
\end{keywords}

\section{Introduction}

Traditionally, in the pre-internet era, advertising was carried out through different advertising channels such as print media, radio, TV, billboards and direct mail. However, in the internet age, with the advent of digital advertising, \emph{multichannel marketing} employs a combination of offline (retail, news papers, bilboards, mail order catalogs, radio etc) and online (websites, display ads, social media, paid search, email, mobile) media in order to better engage with the end-users. Multichannel marketing has posed new challenges in the task of determining the per-channel conversion credits, which is the value of each customer engagement (also called as a customer \emph{touchpoint}) that finally lead to a conversion (buy). Understanding the value of per-channel per-customer touchpoint aids in fair allocation of the budget per-channel leading to acquiring new customers effectively. 

\begin{figure*}[htp]
  \centering
  \subfigure{\includegraphics[scale=0.35]{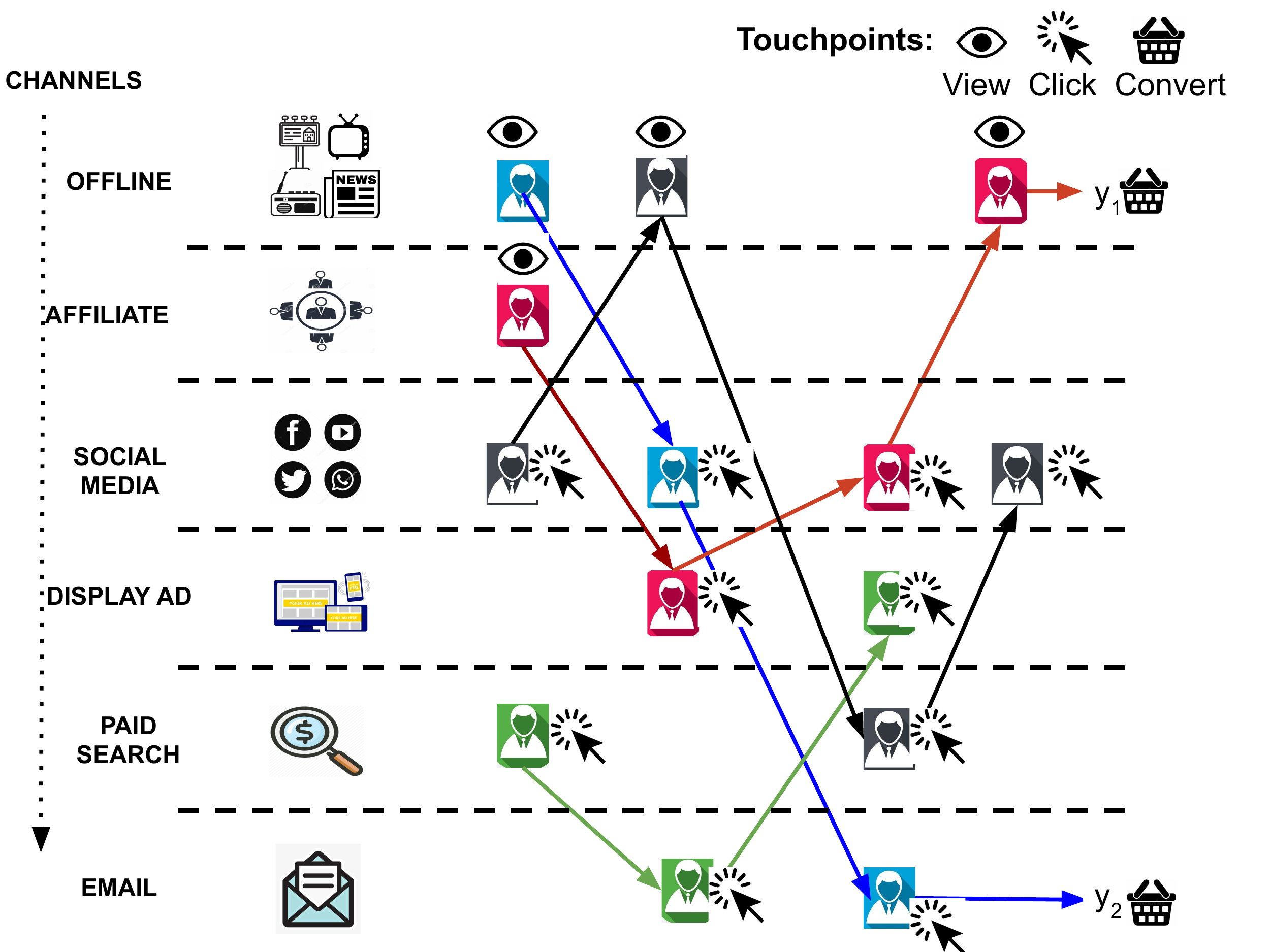}}\quad
  \subfigure{\includegraphics[scale=0.3]{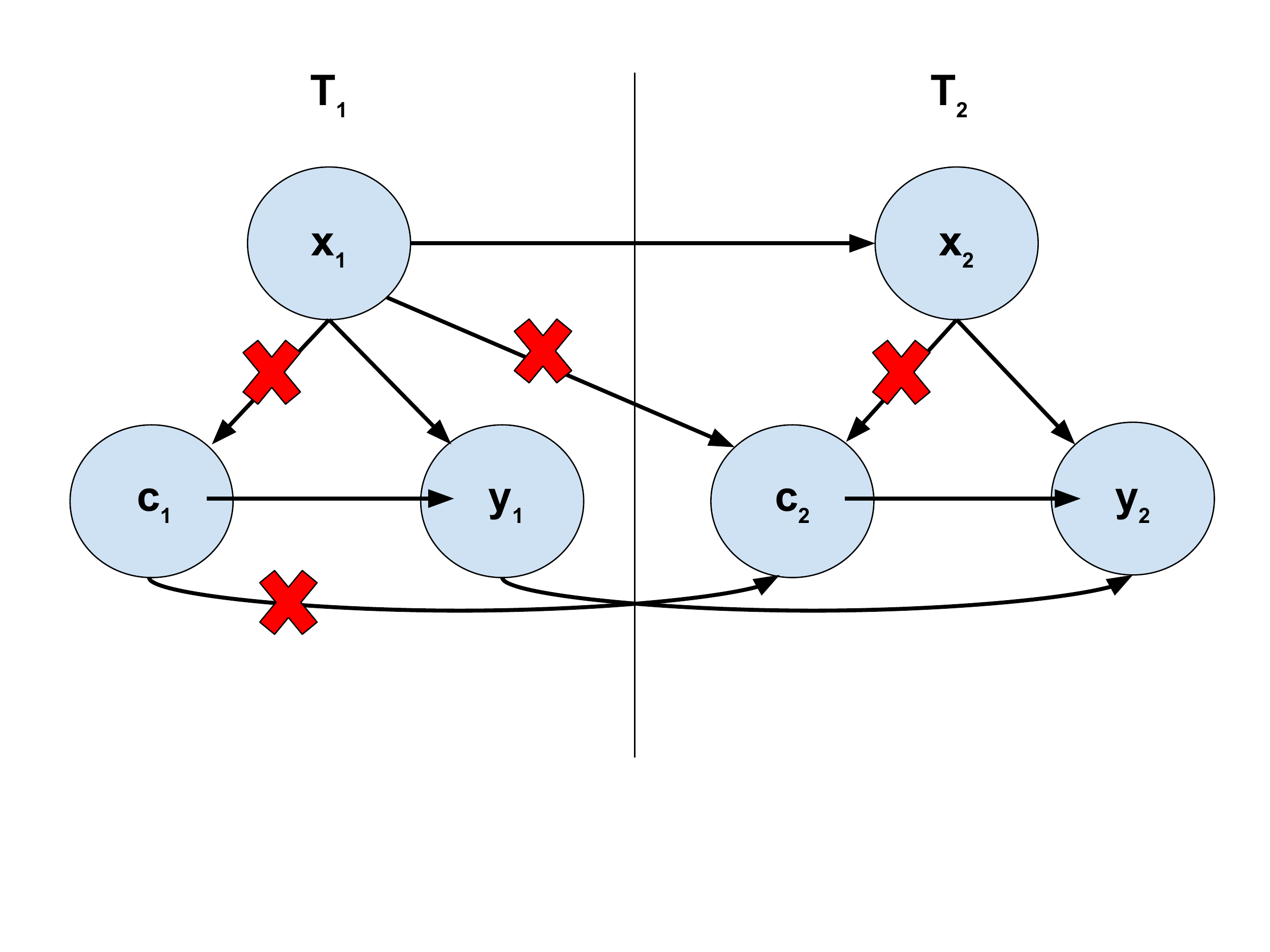}}\\%\label{fig:confounders}\\
  \caption{Figure (a) illustrates typical user journeys across several channels resulting in a temporal sequence of touchpoints. Figure (b) illustrates the temporal confounding effects, where $x_1,x_2$ are the user-context covariates affecting channels $c_1,c_2$ and outcome $y_1,y_2$ at touchpoint $T_1,T_2$ of a user journey. Red-crosses indicate the bias compensating nature of \texttt{CAMTA}}
\label{fig:MTAConfIntro}
\end{figure*}

Multi-touch attribution (MTA) measures the impact of each touchpoint and its contribution towards a conversion, hence determining the value of that specific touchpoint. Data-driven MTA was developed as advertisers started to adopt digital marketing, providing opportunities to incorporate sophisticated and accurate techniques to understand and improve advertising budget allocation. For example, consider a scenario where a user intends to purchase a new laptop. After the customer logs queries on the search-engines, he sees targeted ads from specific laptop manufacturing companies. First the customer sees a display advertisement, which he  ignores. Next, he sees an advertisement on his Instagram feed that catches his attention, and takes him to the laptop manufacturers' website. Finally, the launch of a new product and a promotional offer via email with a discount code leads to a conversion. Such user journeys through several advertisement channels is also depicted in Fig.\ref{fig:MTAConfIntro} (a). In the process, the user-level data such as gender, age, geography, user-level events (clicks, impressions) etc are logged. Conventional techniques such as first-touch attribution, last-touch attribution, linear attribution, etc use such data to measure MTA. These techniques aggregate across users and do not utilize the user-level data for MTA. In order to provide a user-personalised experience by exploiting vast amounts of data, it becomes necessary to design more sophisticated data-driven machine learning techniques. 

Several data-driven approaches are proposed in literature. In \cite{ShaoLogistic} the authors propose a logistic regression method to predict the conversion rate with respect to advertisement occurrences. The authors in \cite{Zhangsurvival} propose data-driven MTA with survival theory, but do not personalise MTA since they neglect user characteristics. Furthermore, \cite{AMTA} use a model-based survival technique where they assume that the impact of ad exposures is additive and fades with time, and employ hazard rate to reflect the influence of an ad exposure. However, the above mentioned techniques do not employ data for user-personalised attribution.

More recently, deep neural network based approaches have been proposed. In \cite{DARNN}, the authors propose a sequential user behavior, sequence learning model, where they learn the attribution from final conversion estimation. An additional modeling constraint in the online advertising scenario is that a single user is exposed to multiple channels, and hence, the data can be interpreted as being \emph{longitudinal} in nature. In \cite{incrementalMTA}, the authors consider multiple touchpoints as different time-steps and employ RNNs to address MTA. They propose an user-level model for purchase of a brand’s product as a function of the user's exposure to ads followed by a fitted model to allocate the incremental attribution. In \cite{DNAMTA}, the authors propose an LSTM based sequential model in conjunction with the attention mechanism to capture the contextual dependency of the touchpoints. An effective approach towards MTA is employing causal inference methods to provide interpretability to a conversion. In  \cite{dalessandro2012causally}, the problem of attribution is posed as a causal estimation problem using Shapely values. In \cite{singal2019shapley}, the authors develop an axiomatic framework for MTA in online advertising, by proposing a novel metric for attribution called as counterfactual adjusted Shapley value. A novel interpretable deep learning model, \texttt{DeepMTA}, for online multi-touch attribution is developed by combining deep learning and cooperative game theory. 

A counterfactual analysis in the context of multi-channel attribution is proposed in \cite{dalessandro2012causally}, where the impact of a channel on user conversion is measured by obtaining the difference in conversion outcomes on a user when he is exposed to the channel as compared to when he is not exposed to the channel. However, since it is not possible to obtain outcomes for both these scenarios for a given user, users are randomly assigned to either group. However, in observational studies pertaining to digital advertising, assigning users to specific channel is not always feasible, and hence, conventional methods to obtain causal parameters for measuring MTA are biased. The main issue in observational data is the presence of confounding (also referred to as \emph{selection bias}), i.e., channel assignment per user depends on user context and hence, users per channel is not random. In order to abate the effect of confounding, several statistical approaches such as sub-classification, weighting, imputations, and propensity score (PS) matching for unbiased per-individual causal estimates have been proposed. Furthermore, modern deep neural network (DNN) based approaches minimize the discrepancy between the distribution of individuals receiving different treatments, in order to emulate a randomized trial \cite{Johannson2018,sharma2020multimbnn}.  

Causal inference on longitudinal observational data provides us an opportunity to understand how user-behavior and buying patterns evolve as a cause-effect relationship under different channels and ad exposures, thus leading to new tools for digital advertising. In the context of time-varying confounding, estimating the effects of time-varying channel exposures are based on marginal Structural Models (MSMs) \cite{MSMRobins,mansournia} and inverse probability of treatment weighting (IPTW). Recently, \cite{CRN} proposed the counterfactual recurrent network (CRN), which integrates domain adversarial training into sequence-to-sequence architecture for estimating treatment effects over time. Furthermore, CRN constructs treatment invariant representations at each time-step, thus avoiding the association between patient history and treatment assignment. However, when applied to digital advertising, CRN by itself does not provide information regarding per-channel conversion credits or attribution. 

In this work, we propose a causal attribution mechanism where we obtain user-personalised MTA for observational data. In particular, we propose the novel DNN architecture, which we refer to as \texttt{CAMTA}. The proposed architecture is based on  CRN \cite{CRN}, with additional functionality to obtain per-channel attribution by employing an attention layer. Our contributions are as follows: 
\begin{itemize}
    \item The key contribution of \texttt{CAMTA} is the use of causal recurrent network architecture for MTA, which helps to compensate for time-varying confounders (representative causes of confounders shown by red-crosses in Fig. \ref{fig:MTAConfIntro} (b)) further leading to reduction of \textit{selection bias} in learning touchpoint credits to conversion, i.e, attribution.
    \item We use a hierarchical network design for prediction of outcomes, which help to overcome highly skewed ratios of conversions in comparison to customer touchpoints.
    \item Extensive validation of \texttt{CAMTA} is done in terms of prediction performance, budget allocation and interpreting users' buying behaviour.
\end{itemize} 
While \texttt{CAMTA} entails inherent benefits of counterfactual analysis, which is to identify which part of the observed profits (due to conversions) is attributable to the impact of the an advertisement, modelling the data using recurrent network aids in analysing time-variation in confounders. Furthermore, since a single channel is used per time-step, we design an attention layer to measure attribution in terms of one attention weight per channel. We demonstrate the efficacy of \texttt{CAMTA} on a challenging real-world Criteo dataset. 

\section{Causal Attention for attribution: Preliminaries and Loss Function}
In this section we shall describe the problem of multi-touch attribution in the presence of time-varying and confounding context information. We motivate the need for learning a deconfounded, channel invariant representation of context for each touch-point. We further describe the use of attention network to learn channel attribution and loss functions employed for learning context representation, channel attribution and outcome action (click, conversion).
\subsection{Multi-touch attribution Preliminaries}
We consider $\mathcal{D} = \{u^{n},\{\mathbf{x}_t^{n}, \mathbf{c}_t,z_{t+1}^{n}\}_{t = 1}^{T^{n}},y^n\}_{n=1}^N$ as the user browsing dataset where a given user (customer) $u^n$ interacts with $T^{n}$ touchpoints. Each touchpoint has a context vector $\mathbf{x}_t^{n}$ which consists of features such as user demographics, advertisement information, website, operating system along with channel $\mathbf{c}_t$, binary outcome of click or non-click $z_{t+1}^{n}$, etc. Note that click outcome $z_{t+1}^{n}$ is observed as a part of context co-variates $\mathbf{x}_{t+1}^{n}$. Each touch point is delivered by one of $K$ channels given by $\mathbf{c}_t = [c_t(1),..c_t(k),..,c_t(K)]$ , where each entry $c_t(k)$ is binary , i.e, $c_t(k) \in \{0,1\}$. In a given time-step, each touchpoint can be represented by one channel, and hence, $\mathbf{c}_t$ is a one-hot vector. The sequence of touchpoints lead to conversion($1$) or non-conversion($0$), leading to a binary outcome $y^n$. Several preliminary works in MTA literature do not use click for determining attribution of channels from touchpoints to conversion \cite{DNAMTA}. However, in real-world the conversion rate of a user is very small in comparison to his interaction with online ads. This leads to problem of class imbalance. To abate this effect, we use intermediate outcomes, click, to learn a hierarchical mapping from touchpoints to click and further to conversions to boost the estimation of sparse conversion behaviour in our proposed \texttt{CAMTA} architecture.

In addition to this, we consider multi-touch observation data which has an inherent issue of \textit{selection bias} due to time-varying confounders. Confounders are user contextual features that impact both, outcome and channel preference at each touch-point. For example, channel preference for the current touchpoint is influenced by factors like user's click behaviour, user's demographics, access to channels etc. (refer Fig.~\ref{fig:MTAConfIntro}). These factors along with channel selected for current advertisement display have a bearing on user's current touchpoint click action, thereby accounting for confounding effect at each touchpoint. We address the issue of time-varying confoundedness using representation learning and further learn attribution of each channel in conversion using an attention network as described in the sequel.

\subsection{Causal Recurrent Network}
%employed in order 
In the causal inference literature \cite{Johannson2018} based on representation learning, the crux of the loss function is based on decorrelating context information and treatment assignment for abating selection bias. In this context, data is sequential since touch-points span different time-steps in a sequential fashion. \texttt{CRN} \cite{CRN} decorrelates user's history (context) from his/her channel preferences (treatment) at each touchpoint. For this, a recurrent neural network is employed to learn latent state representation of context at each touchpoint ($\mathbf{s}_{t}^n$):
\begin{equation}
    \mathbf{s}_t^n = f_{t,s}(\mathbf{x}_t^n,\mathbf{x}_{t-1}^n,...\mathbf{x}_1^n,\mathbf{c}_{t-1},...,\mathbf{c}_1, z_t^n,z_{t-1}^n,...,z_2^n)
\end{equation}
This latent state vector ($\mathbf{s}_t^n$) is a $L$ dimensional representation of user's ad interaction journey consisting of channels offered [$\mathbf{c}_{1}$,...,$\mathbf{c}_{t-1}$], user's demographics [$\mathbf{x}_1^n$,...,$\mathbf{x}_{t-1}^n$,$\mathbf{x}_t^n$] and click outcomes [$z_2^n$,..$z_{t-1}^n$,$z_t^n$] till touchpoint $t$ and hence can be termed as latent user history ($\mathbf{s}_t^n$). This latent user history impacts click outcome $z_{t+1}$ and plays a significant role in personalized ad targetting algorithms \cite{klever2008behavioural} for deciding current channel $\mathbf{c}_{t}^n$. This preferential assignment of the channel leads to \textit{selection bias}. If we proceed without compensating for the selection bias, we obtain biased estimates of the click outcome, which further impacts channel attribution and conversion prediction. Unlike conventional, unitary time-step scenarios, here the impact of biased estimation tends to extend to several time-steps, making it necessary to abate the effect of selection bias at each time-step. In order to abate this effect, we learn channel invariant balanced representation $\mathbf{r}_t^n$ using \textit{MinMax} loss such that $\mathbf{\Phi}_t: \mathbf{s}_t^n \rightarrow \mathbf{r}_t^n$, where $\mathbf{\Phi}_t \in \mathbb{R}^{M \times L}$ is a representation that helps to minimize confounding.

\subsection{\textit{MinMax} Loss}

We use two classifiers, channel-assignment classifier ($\mathcal{C}_{t,c}$) and click outcome prediction classifier ($\mathcal{C}_{t,z}$) at each of the touchpoints for the purpose of learning balanced representation $\mathbf{r}_t^n$. Channel-assignment classifier ($\mathcal{C}_{t,c}$) is a 2 layered, multiple layer perceptron (MLP) with $K$ dimensional outcome fed to a softmax layer for learning propensity of each of the $K$ channels. Similarly, $\mathcal{C}_{t,z}$ is click prediction classifier with 2-layered MLP with $1$ dimensional output fed to sigmoid layer for obtaining probability of click at each touchpoint ($\hat{z}_{t+1}^n$).

Since the objective is to obtain an unbiased representation $\mathbf{r}_t^n$ invariant of channel $\mathbf{c}_t$, channel-assignment classifier should learn equi-propensity for each of $K$ channels. Hence, the loss function ($\mathcal{L}_{t,c}$) for channel-assignment classifier should \textit{maximize}.
\begin{equation}
    \mathcal{L}_{t,c}(\mathbf{\Phi}_t) = -\sum_{k = 1}^K c_t(k)\log (\mathcal{C}_{t,c}(\mathbf{r}_t^n)).
\end{equation}
In addition, this representation must be an accurate predictor of click outcome. For this, click-prediction loss $\mathcal{L}_{t,z}$ should \textit{minimize} the loss function given as
\begin{equation}
    \mathcal{L}_{t,z}(\mathbf{\Phi}_t) = -\sum_{k = 1}^K z_{t+1} \log (\hat{z}_{t+1}^n),
\end{equation}
where $\hat{z}_t^n = \mathcal{C}_{t,z}([\mathbf{r}_t,\mathbf{c}_t])$. Hence, we learn a channel-invariant representation which predicts click outcomes accurately at each of the touchpoints using an overall \textit{MinMax} loss :
\begin{equation}
    \label{eq:rep_loss}
    \mathcal{L}_{r}(\mathbf{\Phi},\lambda) = \sum_{t}\mathcal{L}_{t,z}(\mathbf{r}_t^n)-\lambda\mathcal{L}_{t,c}(\mathbf{r}_t^n),
\end{equation}
where $\mathbf{\Phi} = [\mathbf{\Phi}_1,...\mathbf{\Phi}_{T^n}]$, and the value of $\lambda$ is obtained during hyperparameter tuning as discussed in the Appendix \ref{sec:appendixCAMTAspecs}.

\subsection{Attention Mechanism}
In the previous subsection, we obtained click outcome probability which factors in channel invariant representation of user history '$\mathbf{r}_t$' and channel used at current touchpoint '$\mathbf{c}_t$'. In this section, we address the issue of computing attribution of each touchpoint to conversion, represented by $\hat{y}$. 

 We propose to compute the attribution of each touchpoint to conversion using an attention layer subsequent to the representation layer. In order to compute attribution weights, we leverage the hierarchical attention mechanism as proposed in \cite{yang2016hierarchical}. The click outcome of the $t$-th touchpoint of the causal recurrent network given by $\hat{z}_{t+1}^n$ is processed through a one-layer $\textnormal{tanh}$ MLP to obtain the hidden representation $\mathbf{v}_t^n$ of click outcome, channel and representation $\mathbf{r}_t$. This can be mathematically represented as
\begin{equation}
    \mathbf{v}_t^n = \textnormal{tanh}(\mathbf{W}_v\hat{z}_{t+1}^n +\mathbf{b}_v),
\end{equation}
where $\mathbf{W}_v, \mathbf{b}_v$ are the trainable parameters of MLP.
Then we obtain attribution weights as the similarity of hidden representation $\mathbf{v}_t^n$ with trainable touchpoint context vector $\mathbf{u}$, normalized through a softmax function: 
\begin{equation}
    a_t = \frac{\exp((\mathbf{v}_t^n)^T\mathbf{u})}{\sum_t \exp((\mathbf{v}_t^n)^T\mathbf{u})}.
\end{equation}

\subsection{Conversion Estimation}

In order to obtain the predicted conversion output, we first obtain the hidden state representation of entire sequence $h_t$ as the attention weighted sum of touchpoint representation $v_t$, given by
\begin{equation}
    %s_t = \sum_ta_t\mathbb{C}_z([\mathbf{r}_t,\mathbf{c}_t])
    \mathbf{h}_t^n = \sum_ta_t\mathbf{v}_t^n.
\end{equation}
The sequence representation $\mathbf{h}_t^n$ is passed through a $\textnormal{sigmoid}$ one-layer MLP in order to obtain  conversion prediction $\hat{y}^n$, as given by 
\begin{equation}
    \hat{y}^n = \textnormal{sigmoid}(\mathbf{W}_y^T\mathbf{h}_t^n+\mathbf{b}_y),
\end{equation}
where $\mathbf{b}_y$ is a trainable parameter along with $\mathbf{W}_y$. Furthermore, we use \textit{binary cross entropy} loss for conversion prediction of the sequence, i.e., 
\begin{equation}
    \mathcal{L}_y(\mathbf{W}_y,\mathbf{b}_y,\mathbf{W}_v,\mathbf{b}_v,\mathbf{\Phi},\mathbf{u}) = -y^n \log(\hat{y}^n)
\end{equation}

% click prediction at each touchpoint and sequence conversion prediction.
\section{CAMTA: Proposed Architecture}
\begin{figure}[t!]
\includegraphics[scale=0.70]{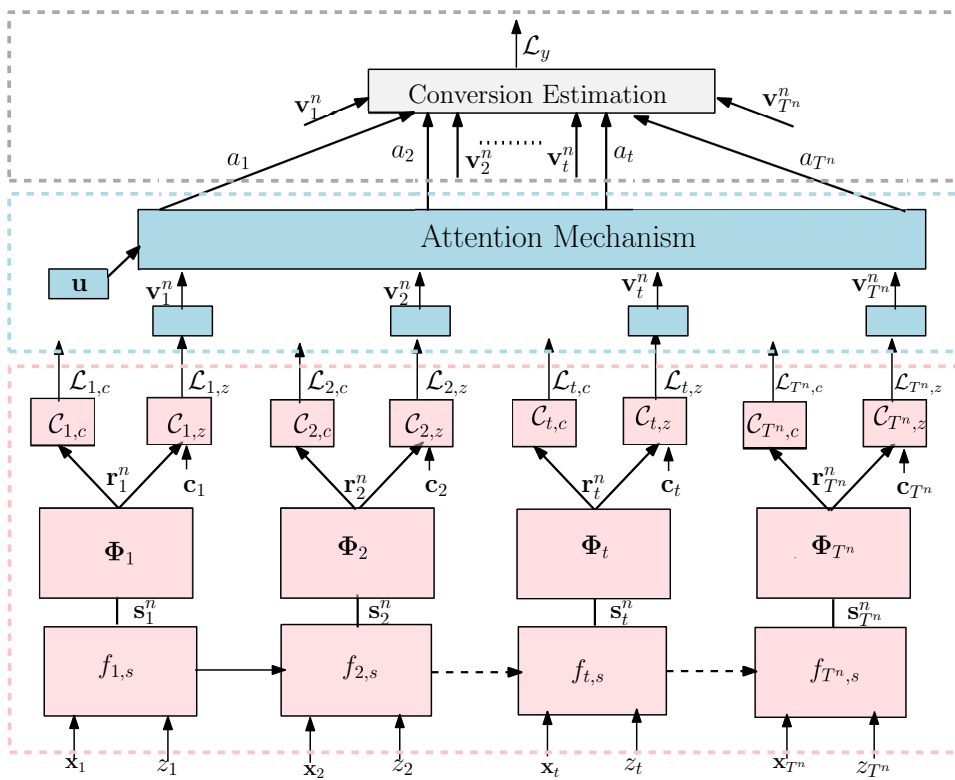}
\caption{Detailed architecture of \texttt{CAMTA}. Pink dashed outline represent Causal Recurrent Network, Blue dashed block show attention mechanism and grey colored blocks stands for conversion estimation component of \texttt{CAMTA}. }
\label{fig:CAMTA}
\end{figure}

In this section we describe the novel \texttt{CAMTA} architecture for multi-touch attribution. The detailed diagram of proposed architecture is shown in Fig. \ref{fig:CAMTA}. The network consists of three major interconnected components, as depicted in Fig.~\ref{fig:CAMTA}. First, the latent state history ($\mathbf{s}_t^n$) of user is learned at each touchpoint which is followed by the reduction of selection bias by employing representation learning using \textit{MinMax} loss eq.~\ref{eq:rep_loss}. This component is termed as \textit{Causal Recurrent Network} and outputs click probability to the subsequent attention layer. Attention  mechanism is used for touchpoint attribution which learns hidden representation $\mathbf{v}_t^n$ of channel, click and context representation and uses similarity between $\mathbf{v}_t^n$ and trainable context vector $\mathbf{u}$ for computing attention of each touchpoint. Finally, a one-layer MLP is employed for conversion prediction. The overall loss function for each sequence is given by:
\begin{equation}
    \label{eq:overall_loss}
    \mathcal{L}(\lambda,\beta) = \mathcal{L}_r(\mathbf{\Phi},\lambda) +\\
    \beta \mathcal{L}_y(\mathbf{W}_y,\mathbf{b}_y,\mathbf{W}_v,\mathbf{b}_v,\mathbf{\Phi},\mathbf{u})
\end{equation}

\section{Experimental Set-up}
In this section, we demonstrate the efficacy of the proposed \texttt{CAMTA} on publicly available real-world live traffic dataset \texttt{Criteo} \cite{DiemertMeynet2017}. First, we discuss the dataset and the processing methodology, followed by evaluation metrics, baseline approaches and implementation specifics of \texttt{CAMTA}. 

\subsection{Dataset \& Data processing}
We use \texttt{Criteo} dataset for validation of the proposed approach. The dataset has more than $16$ million impressions (touchpoints) over $675$ campaigns. Each impression is associated with $9$ categorical covariates, the context of which is masked for confidentiality purposes. Impression logs are also associated with the identity of user and conversion identity. Each user is associated with either single or multiple conversion identities. Hence, we split such user sequences in such a way that each sequence has at most one conversion. Additionally, we omit sequences which consists of more than $20$ touchpoints since those constitute less than $0.5\%$ of all sequences in the data. %Garima, we need to substantiate this. Is 20 touchpoint sequences outliers? Or is it tough to process it? What percentage of the dataset is this. Better add some numbers. 
%%%%%%%%
Each impression consists of information regarding the advertising campaign and total cost incurred. We consider the $675$ advertising campaigns as channels, and randomly select $10$ channels for analysis. Accordingly, we remove sequences which consist of channels other than the selected ones. For the ease of handling and analysing data, we reduce the vocabulary size of covariates by combining categorical features based on the word frequency distribution. We provide the first order statistics of the processed dataset used for evaluation and original \texttt{Criteo} dataset in Table~\ref{tab:FO_stats}. Furthermore, the processed dataset is split into 60:20:20 for training, validation and testing sets.

\begin{table}[]
    %\centering
    \begin{tabular}{|l|l|l|}
         \hline
         & Processed \texttt{Criteo}
         dataset & \texttt{Criteo} dataset \\
         \hline
         No. of users & 44,370 &6,142,256\\
         \hline
         No. of channels & 10 & 675\\
         \hline
         No. of sequences & 46,299&6,755,770\\
         \hline
         No. of touchpoints & 82,590& 16,468,027\\
         \hline
         No. of convert sequence &  2,541&806,196\\
         \hline
         No. of click touchpoints & 27,782&5,947,563\\
         \hline
         
    \end{tabular}
    \caption{First order statistics of processed \texttt{Criteo} dataset}
    \label{tab:FO_stats}
\end{table}

\subsection{Evaluation Metric}
\label{sec:metrics}
We evaluate the proposed approach in two major parts. The first part focuses on evaluating the conversion estimation and click estimation performance in terms for \textbf{log-loss} and area under conversion ROC curve ($AUC$). The log-loss ($\mathbf{LL}$) for conversions is given by:
\begin{equation}
    \mathbf{LL}_{conv} = -\sum_{n = 1}^N y^n log (\hat{y}^n)
\end{equation}
Similarly, log-loss for clicks is given by:
\begin{equation}
    \mathbf{LL}_{click} = -\sum_{n = 1}^N \sum_{t=1}^{T^n} z_t^n \log (\hat{z}_{t}^n)
\end{equation}
The second part focuses on attribution-guided budget allocation performance over historical data \cite{DARNN}. For this, we first compute the \textbf{return on investment} ($ROI$) of each channel:
\begin{equation}
    \label{eq:roi}
    \textnormal{ROI}_k = \frac{\sum_{\forall{y^n = 1}} \sum_ta_t^n\mathbb{I}(c_t(k) = 1)V(y^n)}{\sum_n \sum_t \textnormal{cost}_t^n(k)}
\end{equation}
where $a_t^n(k)$ is the attention weight, $\textnormal{cost}_t^n(k)$ is the monetary expenditure of $k$-th channel, at touchpoint $t$ corresponding to the $n$-th sequence in the data sample, and $\mathbb{I}(\cdot)$ is the indicator function. Subsequently, we allocate budgets across $K$ channels according to 
\begin{equation}
    \label{eq:budget}
    b_k = \frac{\textnormal{ROI}_k}{\sum_{k = 1}^K \textnormal{ROI}_k} * B,
\end{equation}
where $B$ is the total budget to be allocated.
The above budget allocation is intuitive since it is the weighted average of $\textnormal{ROI}$, implying that channels with large $\textnormal{ROI}$ are allotted higher budgets.%Garima, check this interpretation and alter (done)

We then use these re-allocated channel budgets to traverse along testing set impressions ordered by their serving time. If there is no budget left for the channel corresponding to impression, entire sequence is removed from further analysis  and termed as \textit{blacklist} sequence. Current channel's cost is subtracted from the remaining budget of the channel and total number of conversion sequences from non-\textit{blacklist} sequences is the \textit{number of true conversions}. The cost spend on all channels is the \textit{total expenditure} and is used for the computation of attribution-guided marketing budget allocation evaluation. The detailed budget re-allocation algorithm can be found in \cite{DARNN}.
%Back evaluation for budget allocation algorithm as in \cite {} is then used for computing \textit{total cost spend} and \textit{obtained number of conversions} on testing set using budgets as allocated using eq~\ref{eq:roi}, \ref{eq:budget}. 

%In order to evaluate the efficiency of marketing budget allocation using attribution, 
We use cost per action (\textit{CPA}), which is the total expenditure normalised by the number of true conversions, and conversion rate (\textit{CVR}), which is the number of true conversions averaged by number of testing set sequences as metrics for validating budget allocation using attribution obtained from \texttt{CAMTA}.

\subsection{Baseline Models}
\label{sec:baseline}
In this section, we discuss the following baseline approaches for multi-touch attribution against which we compare the proposed \texttt{CAMTA} approach:
\begin{itemize}
\item  Additive Hazard (AH) \cite{Zhangsurvival} is based on survival theory for multi-channel attribution problem in online advertising. It considers both, the impact of different levels of advertising channels and time-decaying effect. 
\item  Logistic regression (LR) approach presented in \cite{ShaoLogistic}, where attributions of each channel is computed using logistic regression.
\item  Additional Multi-touch Attribution (AMTA) \cite{AMTA} which employs mathematical tools from survival analysis. Here, hazard rate is used to measure the influence of online advertisement exposure, based on the assumption that the effect of an advertisement exposure fades with time and the browsing path of users are additive.
\item  Deep Neural Net with Attention Multi-touch Attribution (DNAMTA)  \cite{DNAMTA} is a deep neural network incorporating attention mechanism in addition to an LSTM network. 
%It aims to understand the interactions between online advertising channels and their contributions to customer conversion is a more data-driven way.
\item  Dual-Attention Recurrent Neural Network (DARNN) \cite{DARNN} uses dual-attention RNNs, one for impression-level data and another for clickstream data in order to calculate the effective conversion attribution.
\end{itemize}

We use NVIDIA GK110BL[Tesla K40c] for training and hyperparameter optimisation of all baseline approaches. The details of hyperparameter optimisation for baseline approaches are provided in the Appendix~\ref{sec:appendixBaselinesspecs}.

\subsection{Implementation Specifics}
We obtain an embedding or a continuous representations of each of the categorical covariates prior to the first layer of \texttt{CAMTA}. We learn the weights of embedding layer along with \texttt{CAMTA} weights using loss function as given in  \eqref{eq:overall_loss}. Note that the hyperparameters are selected based on the overall loss function given in \eqref{eq:overall_loss} on the validation dataset. Further, the hyperparameter search space details are given in the Appendix~\ref{sec:appendixCAMTAspecs}.

\section{Experimental Results}

In this section, we demonstrate the experimental analysis of \texttt{CAMTA} on the real-world \texttt{Criteo} dataset. We divide the experimental evaluation into two parts
\begin{itemize}
    \item Analysis of \texttt{CAMTA}'s prediction performance
    \item Analysis of attribution guided budget allocation
\end{itemize}
Subsequently, we analyse the nature attribution weights in convert and non-convert sequences and channel attribution weights based on user's buying behaviour.

\subsection{Prediction performance analysis}

We use $AUC$, $\mathbf{LL}_{conv}$, $\mathbf{LL}_{click}$ as the metric to evaluate prediction performance. We observe that \texttt{CAMTA} achieves high $AUC$ of $0.9591$, and very low values of click and conversion log-loss. An empirical analysis of the baseline approaches discussed in Sec.~\ref{sec:baseline} as compared to \texttt{CAMTA} is as shown in Table~\ref{tab:pred}. We see that \texttt{CAMTA} outperforms the baseline approaches by huge margins. In addition, \texttt{CAMTA} ($\lambda = 0$), i.e., \texttt{CAMTA} network without accounting for confounders under-performs as compared to \texttt{CAMTA}  ($\lambda > 0$), hence highlighting the importance of compensation of selection bias due to confounding context variables. Furthermore, these results also point to the significance of using hierarchical network of impressions-click-conversion prediction which helps to overcome skewed representation of conversions in the data. Note that click is used as an input covariate in all other baselines except for our proposed model \texttt{CAMTA} and \texttt{DARNN} which predicts clicks, and uses it further for prediction of conversions and attribution.

\begin{figure}[]
\centering
\includegraphics[scale=0.475]{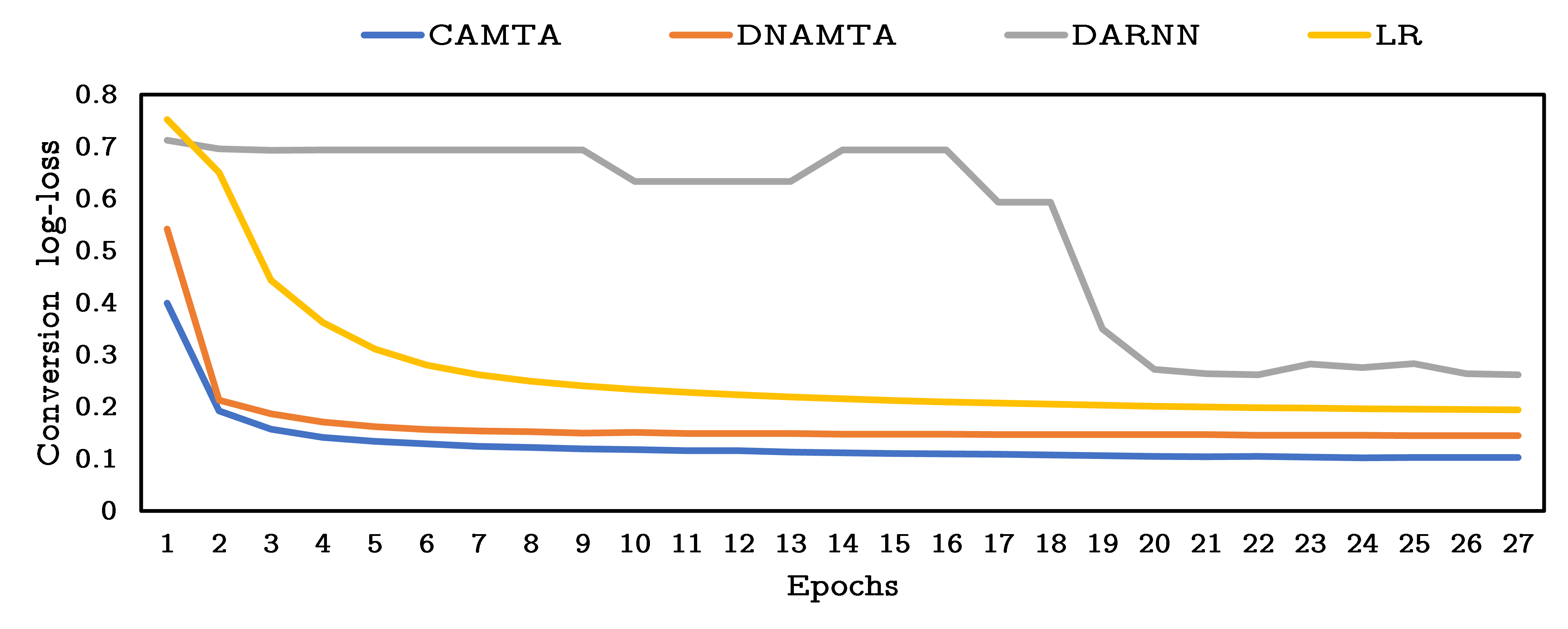}
\caption{$\mathbf{LL}_{conv}$ comparison of \texttt{CAMTA} with baselines}
\label{fig:logloss_conversion}
\end{figure}

\begin{figure}[]
\centering
\includegraphics[scale=0.45]{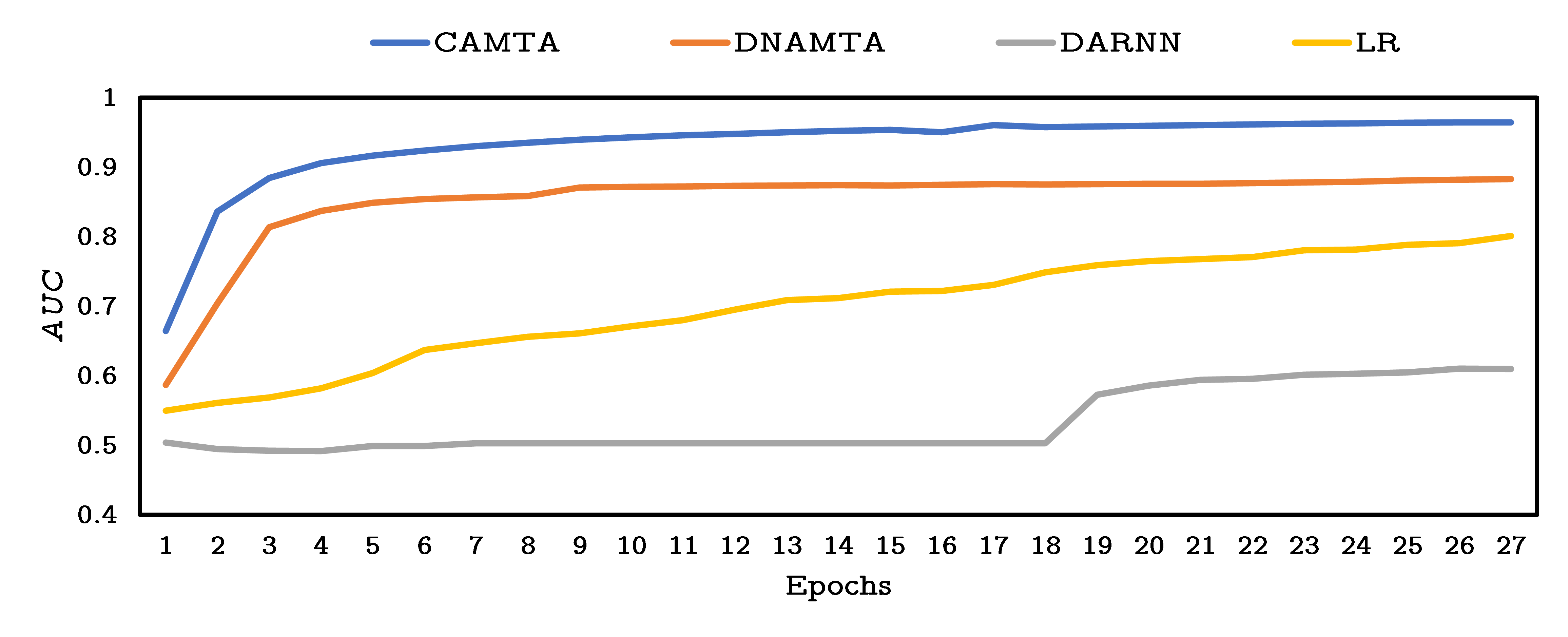}
\caption{Comparison of $AUC$ of \texttt{CAMTA} with the baselines}
\label{fig:AUC_conversion}
\end{figure}

\begin{table}[!h]
    \centering
        \begin{tabular}{ p{2.5cm}p{1.5cm}p{1.5cm}p{1.5cm} }
         \hline
         %\multicolumn{4}{|c|}{Model prediction performance comparison} \\
         %\hline
         Models&$AUC$&$\mathbf{LL}_{conv}$&$\mathbf{LL}_{click}$ \\
         \hline
         \texttt{AH}   & 0.5195  &0.262&  -\\
         \texttt{AMTA}& 0.5309  &0.2418  &-\\
         \texttt{LR} &0.839 & 0.195& -\\
         \texttt{DNAMTA}  &0.9119 & 0.1365&  -\\
         \texttt{DARNN}& 0.6108 &0.2389&0.3195\\
         \texttt{CAMTA} ($\lambda = 0$)& 0.9469 & 0.11 & 0.0785\\
         \texttt{CAMTA}&\textbf{ 0.9591}  & \textbf{0.0977}   &\textbf{0.0639}\\
         
         \hline
        \end{tabular}
    \caption{Comparison of the Prediction performance of \texttt{CAMTA} and baseline schemes}
    \label{tab:pred}
\end{table}
\iffalse
\begin{tabular}{ |p{2.5cm}||p{1.5cm}|p{1.5cm}|p{1.5cm}|  }
 \hline
 \multicolumn{4}{|c|}{Model prediction performance comparison} \\
 \hline
 Models&$AUC$&$\mathbf{LL}_{conv}$&$\mathbf{LL}_{click}$ \\
 \hline
 \texttt{AH}   & 0.5195  &0.262&  -\\
 \texttt{AMTA}& 0.5309  &0.2418  &-\\
 \texttt{LR} &0.839 & 0.195& -\\
 \texttt{DNAMTA}  &0.9119 & 0.1365&  -\\
 \texttt{DARNN}& 0.6108 &0.2389&0.3195\\
 \texttt{CAMTA} ($\lambda = 0$) 0.9469 & 0.11 & 0.0785\\
 \texttt{CAMTA}&\textbf{0.9591}  & \textbf{0.0977}   &\textbf{0.0639}\\
 
 \hline
\end{tabular}
\fi
In Fig.~\ref{fig:logloss_conversion} and Fig.~\ref{fig:AUC_conversion}, we compare the $\mathbf{LL}_{conv}$ and $AUC$ performance of \texttt{CAMTA} with the baseline schemes across epochs. We see that  \texttt{CAMTA} consistently beats state of art approaches by large margins, for different metrics related to conversion prediction.

\subsection{Attribution weights comparison}
%\iffalse
\begin{figure*}[]
\centering
\includegraphics[scale=0.5]{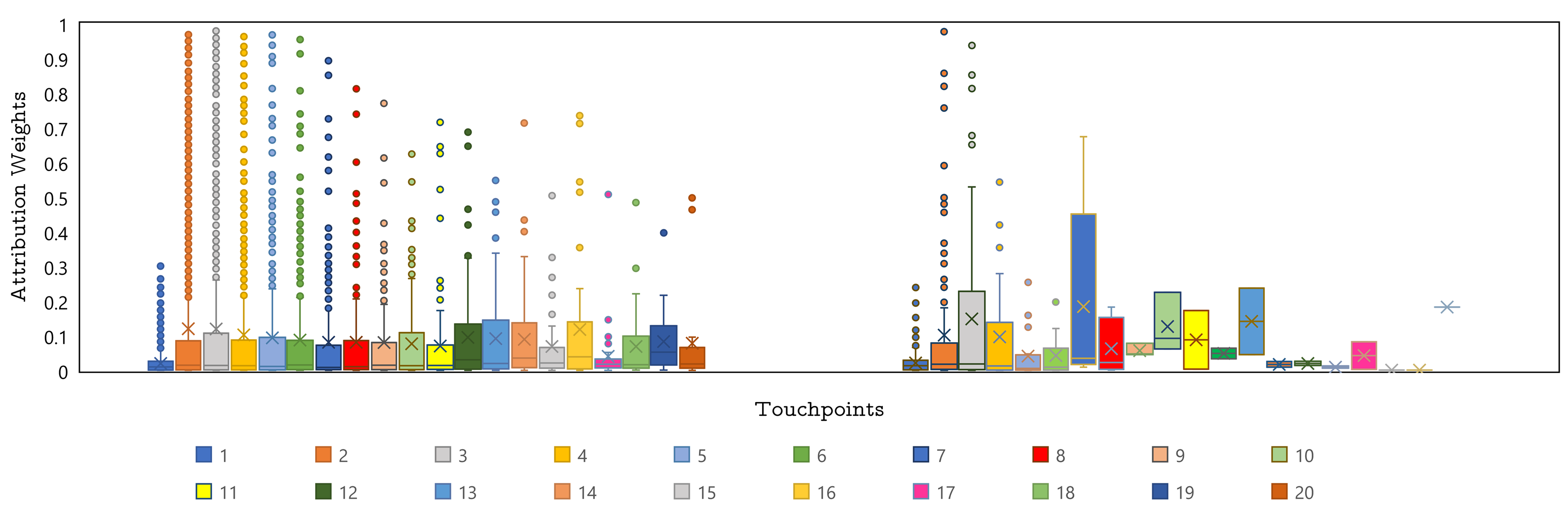}%convert.eps}
\caption{Box plots of attribution weights for non-convert (left) and convert (right) sequences.  }
\label{fig:boxplots}
\end{figure*}
In addition, we employ box plots to analyze the attribution weights of convert and non-convert sequences from testing data for all touchpoints. These box plots are shown in Fig.~\ref{fig:boxplots} where each unique color represents a touchpoint on x-axis, while the y-axis represents the attribution weight for each touchpoint. Box-plots are a standard method to depict the data distribution based on five numbers: quartile $1$, quartile $2$ (median), quartile $3$, minimum, maximum. Minimum and maximum refers to quartile $1$ $\pm$ $1.5$ * (quartile $3$ - quartile $1$). The dots are the outliers beyond maximum, minimum value of attribution weight for each touchpoint. 

From Fig.~\ref{fig:boxplots}, we see that the median attribution weights of non-convert sequences is very close to 0 and distribution across touchpoints is uniform with quartile $3$ for all touchpoints $< 0.15$. However, attribution weights of convert sequences has non-uniform distribution across touchpoints, with touchpoint $3,7,10,13$ having high values of quartile $3$ and touchpoints $7,9,10,12,13$ having high values for quartile $1,2$, clearly pointing to discriminative capability of \texttt{CAMTA} for attribution weights of convert and non-convert sequences.

\subsection{Attribution guided budget allocation}
 We analyze \texttt{CAMTA} for attribution guided budget allocation. Here, we use different proportions of total testing set budget to evaluate the performance of budget re-allocation algorithm as discussed in subsection \ref{sec:metrics}. We use value $V(y^n) = 1$ for computing $ROI_k$ (eq. \ref{eq:roi}) in our implementation. Note that the cost values are scaled to a very small value in \texttt{Criteo} dataset. To highlight the difference of CPA across baselines, we scale \texttt{Criteo} data cost by $1000$. It is seen that if we use budget allocation guided by \texttt{CAMTA}'s attribution, \textit{CPA} is least and \textit{CVR}, which is the conversion rate, is highest for \texttt{CAMTA} for $0.4,0.6,0.8,1$ proportions of total budget. Hence, for attribution-guided budget allocation, our approach outperforms baseline approaches for higher budgets. For smaller budget, \cite{DNAMTA} and our approach are equivalent.
 
It is necessary to be pointed out that this methodology of evaluating budget allocation is an approximate technique in the absence of real-time environment for simulating user, context, channel, cost, outcome. In addition, \texttt{CAMTA} is trained for learning best attribution weights for the prediction of (non-)converts and (non-)clicks. 
%We shall improve upon budget allocation and \textit{CPA} by using these attribution weights for constrained budget optimization as future extension to this work. (I will include this in Discussions. 

\begin{table*}[]
    \tiny
    \centering
        \begin{tabular}{ |p{1cm}|p{0.8cm}|p{0.8cm}|p{0.8cm}|p{0.8cm}|p{0.8cm}|p{0.8cm}| p{0.8cm}|p{0.8cm}|p{0.8cm}|p{0.8cm}| p{0.5cm} |p{0.5cm} |p{0.5cm} |p{0.5cm}|p{0.5cm}| }
         %\hline
         %\multicolumn{13}{|c|}{Budget allocation evaluation results} \\
         \hline
         \multicolumn{1}{|c|}{} &\multicolumn{5}{|c|}{\textit{CPA}} &\multicolumn{5}{|c|}{\textit{CVR}} &\multicolumn{5}{|c|}{\textit{Number of true conversions}} \\
         \hline
         Budget&1&0.8 &0.6&0.4&0.2&1&0.8 &0.6&0.4&0.2&1&0.8 &0.6&0.4&0.2\\
         \hline
         \texttt{AH}  &49.92& 48.7	&48.2&	68.4	&51.3&0.0315&	0.0314&	0.0319&	0.0209&	0.0252	&28&25&	19&	9&	5\\
         \texttt{AMTA}& 53.6&49.9	&54.8&	68.7&	68.5&0.0305&	0.0319&	0.029&	0.0215&	0.0198&30&	28&	19&	10&	4\\
         \texttt{LR} &74.2&72.7	&72.5&	54.2&	36.8&0.0334&	0.0217&	0.0217&	0.0262&	0.0364&19&	13&	11&	9&	5\\
         \texttt{DARNN}  &32.3&31.32&	29.63&	34.91&	29.96&0.0514&	0.0522&	0.0535&	0.0453&	0.0576&66&	62&	59&	29&	9\\
         \texttt{DNAMTA}& 9.59&9.58&9.53	&9.54	&\textbf{14.31}	&0.1346&0.1233	&0.1029	&0.0803	&\textbf{0.0422}	&186&163	&130	&88	&\textbf{19}\\
         \texttt{CAMTA}&\textbf{9.51}&\textbf{9.56}	&\textbf{9.44}	&\textbf{9.41}&	16.04&	\textbf{0.1358}&\textbf{0.124}&	\textbf{0.1042}&\textbf{0.0814}&	0.0374&\textbf{190}&	\textbf{170}&	\textbf{138}&	\textbf{89}&	17\\
         \hline
        \end{tabular}
    \caption{Comparative  analysis of \textit{CPA}, \textit{CVR}, \textit{Number of true conversions} for $0.2,0.4,0.6,0.8,1$ proportion of total budget }
    \label{tab:budget_allocation}
\end{table*}

\iffalse
\begin{tabular}{ |p{1.4cm}||p{1cm}|p{1cm}|p{1cm}| p{1cm}|p{1cm}|p{1cm}|p{1cm}| p{1cm}|p{1cm}|p{1cm}|p{1cm}| |p{1cm}| }
 \hline
 \multicolumn{13}{|c|}{Budget allocation evaluation results} \\
 \hline
 \multicolumn{1}{|c|}{} &\multicolumn{4}{|c|}{CPA} &\multicolumn{4}{|c|}{CVR} &\multicolumn{4}{|c|}{Conversion Number} \\
 \hline
 Budget&0.8 &0.6&0.4&0.2&0.8 &0.6&0.4&0.2&0.8 &0.6&0.4&0.2\\
 \hline
 \texttt{AH}  & 48.7	&48.2&	68.4	&51.3&	0.0314&	0.0319&	0.0209&	0.0252	&25&	19&	9&	5\\
 \texttt{AMTA}& 49.9	&54.8&	68.7&	68.5&	0.0319&	0.029&	0.0215&	0.0198&	28&	19&	10&	4\\
 \texttt{LR} &72.7	&72.5&	54.2&	36.8&	0.0217&	0.0217&	0.0262&	0.0364&	13&	11&	9&	5\\
 \texttt{DARNN}  &31.32&	29.63&	34.91&	29.96&	0.0522&	0.0535&	0.0453&	0.0576&	62&	59&	29&	9\\
 \texttt{DNAMTA}& 9.58&9.53	&9.54	&\textbf{14.31}	&0.1233	&0.1029	&0.0803	&\textbf{0.0422}	&163	&130	&88	&\textbf{19}\\
 \texttt{CAMTA}&\textbf{9.56}	&\textbf{9.44}	&\textbf{9.41}&	16.04&	\textbf{0.124}&	\textbf{0.1042}&\textbf{0.0814}&	0.0374&	\textbf{170}&	\textbf{138}&	\textbf{89}&	17\\
 \hline
\end{tabular}
\fi

\subsection{User-Behaviour using attribution}
In this subsection, we analyze channel attribution weights for different types of users categorized based on their \textit{return} to advertiser. First, we compute return of each impression $ret_t = \frac{a_t^n\hat{y}^n}{cost_t^n}$, where $a_t^n$ is the attention weight of each impression, $cost_t^n$ is the money spend for the impression, $\hat{y}^n$ is the probability of conversion as obtained using \texttt{CAMTA}. We compute the average of $ret_t$ across users' touchpoints in order to obtain users' \textit{return} to advertiser. Based on this, users are categorized as low-, medium-, high- \textit{return} users (using $3$-means clustering) constituting of $93.26\%$,$5.88\%$ and $0.90\%$ of testing set user population. 

\begin{figure*}[]
\centering
\includegraphics[scale=0.6]{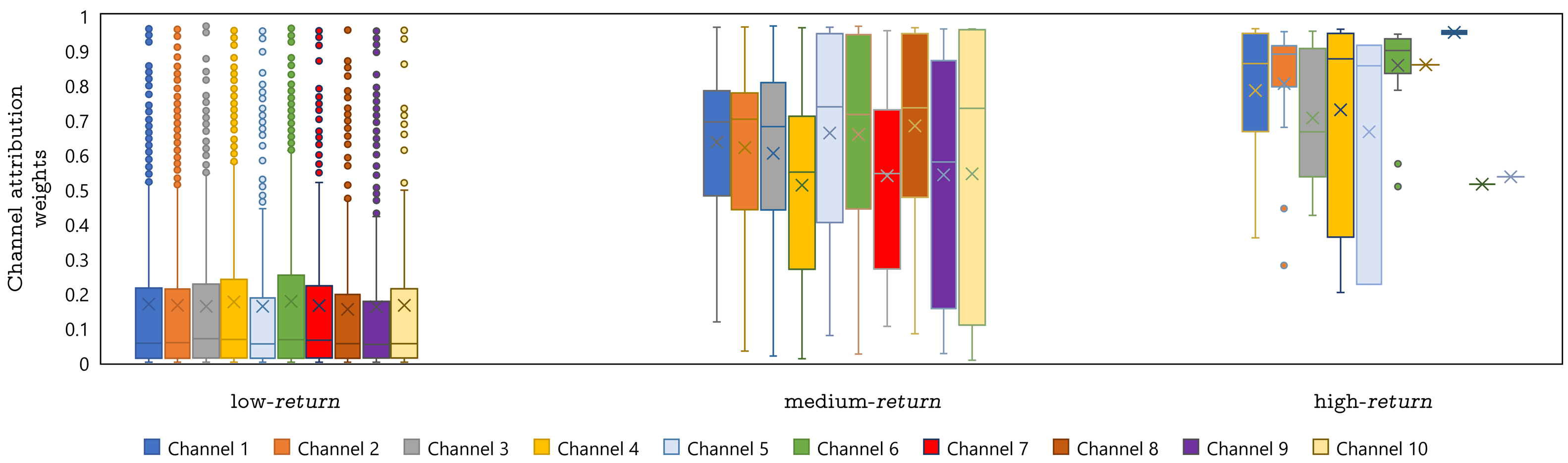}%cluster.eps}
\caption{Channel attribution box plots for low-, medium-, high- \textit{return} user-groups shown in left, middle and right subplots respectively.}
\label{fig:user}
\end{figure*}

 Further, we analyze the user level weighted average of attention (attribution) weights for each of the $K$ channels using box-plots for low-, medium-, high- \textit{return} user-groups. It can be observed from Fig.~\ref{fig:user} that, for low-\textit{return} user-group, the channel attribution weights are consistently low for all channels. Hence, we can conclude that low- \textit{return} users do not have high affinity for any of the channels. Medium- and high- \textit{return} users have high propensity for all channels (fig. \ref{fig:user}) while high-\textit{return} users are particularly more prone to buying using channel $1$ to $6$. 

\section{Discussion and Conclusion}

In this work, we designed \texttt{CAMTA}, a DNN based MTA framework by employing a causal recurrent network for abating selection bias, followed by an attention layer for computing the attribution weights, and a single layer MLP for predicting conversions. In the context of the challenging real-world Criteo dataset, we showed that \texttt{CAMTA} is able to outperform several state-of-the-art baselines such as \texttt{DARNN}, \texttt{DNAMATA}, \texttt{AMTA}, etc in terms of prediction accuracy (Log-loss and AUC). We have also presented experiments towards interpreting the per-channel attribution weights. Using box plots we observed that median attribution weights of non-convert sequences are very  close  to  $0$  and  distribution of touchpoints  is  uniform,   while attribution  weights  of  convert sequences  has  non-uniform  distribution  across touchpoints,  highlighting the discriminative  power of \texttt{CAMTA} with respect to convert  and  non-convert  sequences. Furthermore, we gave insights into budget allocation using the attribution weights obtained using \texttt{CAMTA}. Perhaps one of the most notable achievement of \texttt{CAMTA} is user behavior modelling. Here, we showed that there is considerable behavioral differences observable between low-\textit{return} and medium/high-return user-groups, as measured using the channel attribution weights obtained using \texttt{CAMTA}. While attribution weights remained low and uniform acorss all channels for low- \textit{return} users, indicating low affinity for any of the channels, medium- and high- \textit{return} users displayed high propensity for all channels. Furthermore, high-\textit{return} users were more prone to buying via channel $1$ to $6$.

We conclude that the temporal modeling of the MTA problem, coupled with compensating for selection bias help the attention layer in \texttt{CAMTA} to derive reliable inferences on the buying patterns and user behavior. Furthermore, \texttt{CAMTA} provides an unbiased interpretation angle to the attribution weights obtained from the DNN architecture, which was not available in \cite{DARNN}. As future work, we shall address problems related to constrained budget optimization in digital advertising and axiomatic attribution approach embedded into \texttt{CAMTA}.

\begin{appendices}
\section{\texttt{CAMTA} Hyperparameter Search}
\label{sec:appendixCAMTAspecs}

We train \texttt{CAMTA} using the Adam optimiser for 50 epochs. The hyperparameter search space is given in Table \ref{tab:hyperparameter_CAMTA}

\begin{table}[!b]
    \centering
    \begin{tabular}{p{5.5cm}c}
         \hline
        \textbf{Hyperparameter}&\textbf{Search Range} \\
         \hline
         Learning Rate& 0.01, 0.001, 0.0001\\
         Batch Size&128, 256, 512\\
        Recurrent Neural Network (RNN) Hidden Units&64, 128, 256\\
        Dimension of $\mathbf{r}_t^n$ ($L$) &32, 64, 128\\
        MLP Hidden Units \& Embedding Size &64,128,256\\
        RNN Dropout Probability&0.1, 0.2, 0.3\\
        ($\lambda$,$\beta$)& (1,1),(5,5),(5,10)\\
        \hline
         
    \end{tabular}
    \caption{Hyperparameter search range for the \texttt{CAMTA} model}
    \label{tab:hyperparameter_CAMTA}
\end{table}

\section{Baseline Hyperparameter Selection}
\label{sec:appendixBaselinesspecs}

For baseline approaches, AMTA, DNAMTA and DARNN model hyperparameters are selected according to $\mathbf{LL}_{conv}$ of validation dataset. We train using Adam optimiser for 50 epochs. The hyperparameter search space is given in Table \ref{tab:hyperparameter_baseline}.

%\vspace{5mm} %5mm vertical space

\begin{table}[!h]
    \centering
    \begin{tabular}{p{5.5cm}c }
         
         \hline
        \textbf{Hyperparameter}&\textbf{Search Range} \\
         \hline
         Learning Rate& 0.01, 0.001, 0.0001\\
         Batch Size&128, 256, 512\\
        RNN Hidden Units&64, 128, 256\\
        MLP Hidden Units&64,128,256\\
        RNN Dropout Probability&0.1, 0.2, 0.3\\
        \hline
    \end{tabular}
    \caption{Hyperparameter search range for the baseline approaches}
    \label{tab:hyperparameter_baseline}
\end{table}
\iffalse
%\vspace{5mm} %5mm vertical space
\begin{tabular}{ |p{4.5cm}|p{4.5cm}|  }
 \hline
\textbf{Hyperparameter}&\textbf{Search Range} \\
 \hline
 Learning Rate& 0.01, 0.001, 0.0001\\
 Batch Size&128, 256, 512\\
Recurrent Neural Network (RNN) Hidden Units&64, 128, 256\\
Multi-Layer Perceptron Hidden Units&64,128,256\\
RNN Dropout Probability&0.1, 0.2, 0.3\\
\hline
\end{tabular}
\fi
\end{appendices}

%\newpage

%\appendix
%\section*{Appendix A.}
%\label{app:theorem}

% Note: in this sample, the section number is hard-coded in. Following
% proper LaTeX conventions, it should properly be coded as a reference:

%In this appendix we prove the following theorem from
%Section~\ref{sec:textree-generalization}:
%\bibliographystyle{natbib}

\bibliography{sample}

\end{document}